\pdfoutput=1

\documentclass[11pt]{article}

\usepackage[]{EACL2023}

\usepackage{times}
\usepackage{latexsym}

\usepackage[T1]{fontenc}

\usepackage[utf8]{inputenc}

\usepackage{microtype}

\usepackage{inconsolata}

\usepackage{graphicx}

\usepackage{booktabs} 
\usepackage{makecell} 
\usepackage{multirow} 
\usepackage{stfloats}
\usepackage{float}
\usepackage{amsmath}
\usepackage{tablefootnote}
\title{Transparent Semantic Change Detection with Dependency-Based Profiles}

\author{
 \textbf{Bach Phan-Tat\textsuperscript{1}},
 \textbf{Kris Heylen\textsuperscript{1}},
 \textbf{Dirk Geeraerts\textsuperscript{1}},
 \textbf{Stefano De Pascale\textsuperscript{1}},
 \textbf{Dirk Speelman\textsuperscript{1}}
\\[0.5ex]
 \textsuperscript{1}Department of Linguistics, KU Leuven
\\[0.5ex]
 \small{
   \textbf{Correspondence:} \href{mailto:ttbach.phan@kuleuven.be}{ttbach.phan@kuleuven.be}
 }
}

\begin{document}
\maketitle
\begin{abstract}
Most modern computational approaches to lexical semantic change detection (LSC) rely on embedding-based distributional word representations with neural networks. Despite the strong performance on LSC benchmarks, they are often opaque. We investigate an alternative method which relies purely on dependency co-occurrence patterns of words. We demonstrate that it is effective for semantic change detection and even outperforms a number of distributional semantic models. We provide an in-depth quantitative and qualitative analysis of the predictions, showing that they are plausible and interpretable.
\end{abstract}

\section{Introduction}

\begin{figure}[t!]
    \centering
    \includegraphics[width=\linewidth]{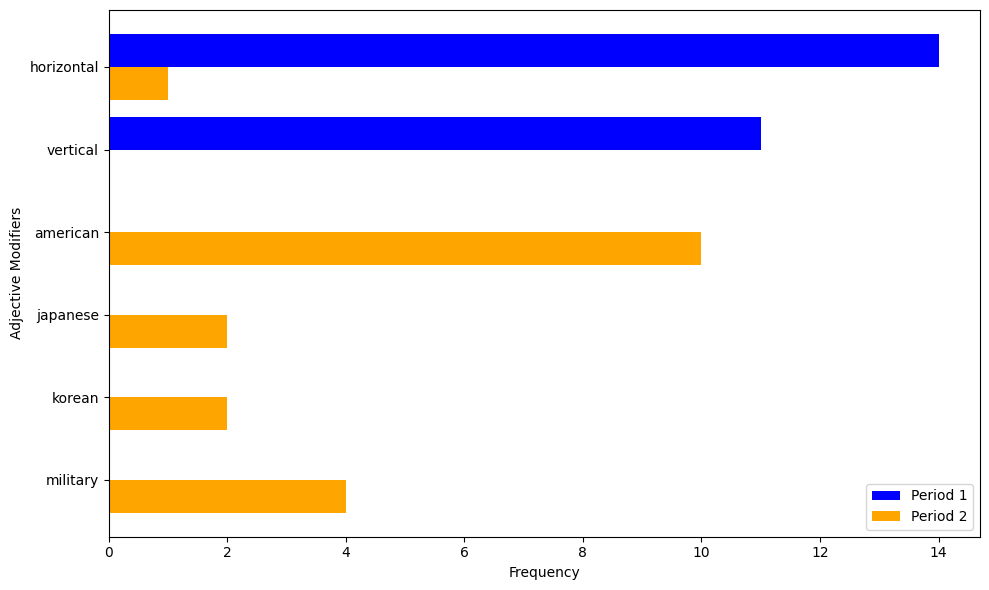}
    \caption{Changes in the frequencies of 6 adjectival modifiers of the English noun \textit{plane} between the 2 sub-corpora for English of the SemEval 2020 shared task 1 \cite{schlechtweg_semeval-2020_2020}. \textit{Plane} is annotated as semantically changed in the dataset}
    \label{fig:plane}
\end{figure}

Lexical semantic change detection is a well-established field in Natural Language Processing (NLP), with several shared tasks for different languages, amongst which the most popular one is SemEval 2020 Task 1 \cite{schlechtweg_semeval-2020_2020}. Most of the models use either static word embeddings like word2vec \cite{mikolov_efficient_2013} or ELMo \cite{peters_deep_2018} or contextualised ones such as BERT \cite{devlin_bert_2019} or S-BERT \cite{reimers_sentence-bert_2019} which is the current state-of-the-art \cite{cassotti_xl-lexeme_2023}. Despite the differences in their architectures, they all build upon the distributional semantics hypothesis \cite{firth_synopsis_1957} and all indirectly represent meanings as numerical vectors, abstracting away from the level of individual co-occurring items. Yet, linguists have long observed that we can study semantics and word-usage by directly looking at the untransformed co-occurrence contexts of the target \cite{mcenery_usage_2019, stubbs_collocations_1995}, which suggests that we could study semantic change by looking at shifts in the contexts of the target. However, most co-occurrence methods, including a recent one by \citet{mcenery_usage_2019}, rely on surface co-occurrences (i.e., bag-of-words approaches to co-occurrence). Surface co-occurrence often suffers from accidental and/or indirect co-occurrences and the arbitrary choice of the span size \cite{evert_corpora_2008}. Syntactic (dependency) co-occurrence \cite{evert_corpora_2008, garcia_method_2019, seretan_syntax-based_2011}, with its different dependency relations, is more helpful to separate the signals. Consider the English noun \textit{plane}: in the 20th century, the AIRCRAFT meaning emerged and became more dominant with the rise of such adjectival modifiers as \textit{american}, \textit{korean}, \textit{japanese}, \textit{military} whereas the DIMENSIONAL meaning became less dominant with the decrease of such adjectival modifiers as \textit{horizontal}, \textit{vertical}, as shown in Figure \ref{fig:plane}.

Using distributional shifts of the dependency co-occurrences to detect lexical semantic change is the focus of this paper.  We investigate to what extent we could use explicit linguistic information to detect semantic change. Our main hypothesis is that the degree of semantic change of a word is reflected in changes in its dependency co-occurrences. Note that we are not trying to set a new state-of-the-art but to show the usefulness of explicit linguistic information and more simple, transparent techniques.

We investigate this question experimentally using the dataset for English, German, Swedish, Latin from SemEval 2020 Task 1. Our main findings are the following:
\begin{enumerate}
    \item Tracing the changes in the distribution of dependency co-occurrences outperforms a number of distributional models.
    \item As our method is transparent by design, we can easily understand how a word has changed (as in the \textit{plane} example above). In contrast, dense embedding models often require post-hoc interpretability methods or additional analyses (e.g., Integrated Gradients \cite{Sundararajan2017AxiomaticAF}, LIME \cite{ribeiro-etal-2016-trust} or SHAP \cite{Lundberg2017AUA},  see surveys in \citealp{Madsen_2022,arrieta2019explainableartificialintelligencexai})  to obtain comparable, feature-level explanations. This makes our approach particularly convenient for linguistic and humanities studies where the explanation needs to be directly inspectable.
    \item Our approach is compatible with linguistic theories of lexical semantics and semantic change \citep[e.g.,][]{geeraerts_lexical_2023, geeraerts_diachronic_1997, hanks_lexical_2013, dalpanagioti_corpus_2019}, allowing researchers to bridge the gap between NLP and diachronic linguistics.
\end{enumerate}

\section{Related work}

A long tradition in distributional semantic models treat lexical meaning as a function of usage: words are characterised by the contexts they appear in, and semantic change is operationalised as a shift in these contextual distributions over time. Within this tradition, work roughly splits into two dominant methodological styles. 

The first tracks change directly in interpretable co-occurrence evidence (e.g., shifts in salient collocates or in the distribution of context features). This line has strong roots in mid-20th-century structuralism, combining Harris’s distributional analysis \cite{harris_distributional_1954} with Firth’s contextual theory of meaning \cite{firth_synopsis_1957}. With the rise of corpus linguistics and concordancing, this idea became a practical methodology. Corpus linguists compared a target word’s prominent collocates and recurring patterns across corpora and later across time slices, often supported by association measures for identifying characteristic collocations \cite{church_word_1990, sinclair_corpus_1991}. This collocational tradition also developed notions such as semantic preference/prosody, which are systematic attitudinal or functional tendencies inferred from a word’s typical collocational environments \cite{partington_patterns_1998, stubbs_collocations_1995}. This naturally lends itself to diachronic analysis when those environments shift. This is done by building time-sliced collocation profiles and examining how association-ranked collocates emerge, fade, or are replaced across periods, keeping the evidence directly inspectable. Case studies have used such collocation turnover to document long-term changes in conventionalised combinations \citep[e.g.,][]{alba-salas_life_2007, pettersson-traba_measuring_2021}. Some of the more recent semi-automatic methods were developed by  \citet{garcia_method_2019} and \citet{mcenery_usage_2019}. Different tools have also made this style scalable, enabling extraction and comparison of diachronic collocation phenomena in large corpora such as AntConc \cite{anthony_antconc_2014}, \#Lancsbox \cite{LancsBox_v6}, SketchEngine \cite{kilgarriff_sketch_2014}, DiaCollo \cite{jurish_diacollo_2015, jurish_using_2020}. 

The second maps co-occurrence evidence into vector-space representations (count-based or neural embeddings). The first models used an untransformed co-occurrence matrix (later changed to other measures such as Mutual Information (MI) or Log Likelihood Ratio (LLR)), with each row corresponding to a word and the columns words in the vocabulary \cite{bullinaria_extracting_2007, gulordava_distributional_2011, schutze_automatic_1998}. The result is that each word will be represented by a sparse count vector. Predictive neural embedding architectures like word2vec \cite{mikolov_efficient_2013}, GloVe \cite{pennington_glove_2014}, FastText \cite{bojanowski_enriching_2017} could be said to capture similar information as the count-based approach but are more powerful at abstracting information from the corpus with denser vectors \cite{baroni_dont_2014}. Contextualised models such as BERT \cite{devlin_bert_2019} are even more powerful in the sense that they can distinguish between individual instances of word usages instead of representing one word as one vector \cite{tahmasebi_computational_2023}. Several attempts have been made to integrate explicit linguistic information into neural network models, such as Dependency Skip-gram \cite{levy-goldberg-2014-dependency} or SynGCN and SemGCN \cite{vashishth-etal-2019-incorporating}. Semantic similarities between words or tokens (in the case of contextualised models) are often calculated using cosine distance between their vectors. For semantic change detection tasks, the standard workflow is as followed \cite{tahmasebi_computational_2023, tahmasebi_survey_2021}: 

\begin{enumerate}
    \item Data are split into different periods;
    \item Semantic vectors of the target words are trained for each time period,;
    \item Different embedding spaces are aligned;
    \item Semantic changes are quantified by comparing aligned representations, often via cosine distance between word vectors, or via clustering comparison of representations across periods.
\end{enumerate}

While collocation-profile approaches keep the evidence directly interpretable, they often yield descriptive outputs that are harder to compare systematically across datasets and languages. As lexical semantic change (LSC) research moved toward standardised evaluation with different standard shared tasks and dataset, amongst which the most popular one is SemEval 2020 Task 1 \cite{schlechtweg_semeval-2020_2020}, methods that produce compact numeric representations and a single comparable distance score became increasingly attractive. As a result, the majority of works and even the current state-of-the-art are all applications of this line of method \citep[e.g.,][]{cassotti_xl-lexeme_2023, geeraerts_lexical_2023, giulianelli_analysing_2020, kim_temporal_2014, kulkarni_statistically_2015, kutuzov_diachronic_2018, montariol_scalable_2021, tahmasebi_computational_2023, tahmasebi_survey_2021}.

There are a few exceptions to this trend, who did not rely on dense numerical vectors, such as \citet{ryzhova_detection_2021} and \citet{kutuzov_grammatical_2021} who used the distributions of grammatical features to detect semantic changes and achieved great results, even surpassing many embedding methods in some tasks and datasets; or \citet{tang-etal-2023-word} who compared sense distributions to detect semantic change. Inspired by their work, we want to investigate the ability of using only explicit linguistic information (in this case, syntactic co-occurrences) in capturing word meaning change. 

Most of the work in collocation/co-occurrence analysis uses bag-of-words window co-occurrence and suffers from arbitrariness in their choice of context windows and from accidental co-occurrences \cite{evert_corpora_2008}. We instead rely on co-occurrence analysis with dependency defined contexts. Rather than using surface window contexts, it treats dependency relation slots (slots) as interpretable dimensions, so predicted change could be attributed to concrete shifts in those dependency environments. This is conceptually similar to the DepDM model of the Distributional Memory Framework \cite{baroni-lenci-2010-distributional} and to the Sketch Engine's Word Sketch Difference \cite{kilgarriff_sketch_2014} but in our case, we keep the raw dependency relations instead of merging them, as will be discussed below. Also, most of them only focus on the salient co-occurrences with such measures as logDice \cite{rychly_lexicographer-friendly_2008}, MI \cite{church_word_1990} or Local MI \cite{baroni-lenci-2010-distributional},  and LLR \cite{dunning_accurate_1993}, thus do not measure the dynamic of the shift of the co-occurrence profile. We quantify semantic change using Jensen-Shannon Divergence \cite{menendez_jensen-shannon_1997}. It is calculated between the distribution of the lexical fillers (slot fillers) of each slot across periods. We evaluate our method in the standard shared tasks, SemEval 2020 Task 1 \cite{schlechtweg_semeval-2020_2020} with 4 languages. Because we use all dependency slots, our method is unsupervised, making it directly comparable to other systems. Finally, we also conduct in-depth qualitative analyses to highlight both the usefulness and limitations of our approach.

\section{Data and tasks}
\label{sec:data and tasks}
Following the SemEval-2020 task on Unsupervised Lexical Semantic Change Detection, we formulate the task as either binary classification or graded ranking, corresponding to Subtask 1 and Subtask 2 in SemEval-2020 \cite{schlechtweg_semeval-2020_2020}. In Subtask 1, given a set of target words, a system has to predict whether each word gained or lost at least one sense between two time periods. In Subtask 2, a system has to rank target words by their degrees of semantic change. Annotating data for semantic change is non-trivial since it requires judging meaning differences across many usage instances from each period. A widely adopted annotation protocol is DURel \cite{schlechtweg_diachronic_2018}, where annotators are asked to compare pairs of sentences containing the target word and judge whether the word is used in the same or a different sense, then these pairwise judgments are aggregated into a final change score, either binary or continuous. This was the procedure used in the creation of the SemEval 2020 dataset. It covers four languages, namely English (37 targets), German (48), Latin (40), and Swedish (32), with manual annotations for both subtasks. Each language is accompanied by a diachronic corpus split into two time periods. Using that dataset allows us to directly compare our approach against state-of-the-art distributional systems under a shared evaluation setup.

\section{Method\footnote{All the codes used in this paper are available at \url{https://github.com/phantatbach/LChange26-Dep}}}

\subsection{Data preprocessing}
Because our method requires dependency relations, which are not available in the original dataset, we first parsed the data using Stanza \cite{qi_stanza_2020} given its high performance on different languages. Upon inspection, we noticed some discrepancies between the Stanza outputs and the lemmatised SemEval data in \textbf{part-of-speech (POS) tags} and \textbf{lemmas}.

\paragraph{POS mismatches} The original dataset did not take into account the POS tags of the target lemmas except for English where targets are provided as \texttt{lemma\_POS} combinations (e.g., \texttt{attack\_nn} but not \texttt{attack\_vb}) because annotation was performed for specific \texttt{lemma-POS} pairs \cite{schlechtweg_semeval-2020_2020}. However, because our package requires the POS tags as part of the input, ensuring POS consistency is crucial. We found cases where tokens corresponding to nominal targets were tagged as \texttt{PROPN} by Stanza (causing missed target occurrences if we filter by \texttt{NOUN}), and cases where adjectival targets were tagged as \texttt{VERB} when used as participles.

\paragraph{Lemma mismatch} We also found cases where Stanza’s lemmatization diverges from the official target lemma due to orthographic variation and OCR errors. One notable example is the German noun \textit{Lyzeum}, which is consistently lemmatised as \textit{Lyceum} by Stanza in the earlier period (old standard spelling), leading to systematic mismatches with the SemEval target form.

To quantify the impact of these discrepancies, we prepared two parallel versions of the parsed data, Stanza-faithful (SZ) and SemEval-faithful (SE). In the SZ version for all languages, we only do minimal normalisation to the target lemmas, namely  changing \texttt{PROPN} to \texttt{NOUN} and spelling variations. In the SE version of German, Latin, Swedish, we ensure that the Stanza results match those of the original lemmatised dataset by replacing all POS tags of the target lemmas with \texttt{TAR} and normalising the spelling variations of the target lemmas. For the English SE version, we split the dataset into multiple sub-datasets based on the official lemmatised files where the targets are marked. Each sub-dataset corresponds to one target lemma.

\subsection{Basic procedure}
\label{sec:basic procedure}
For the main experiment, we used SynFlow \cite{phan_tat_2025_SynFlow}, a custom package developed specifically for this method.

For each target lemma (e.g., the English noun \textit{plane}), we extract all of its dependency slots (e.g., adjective modifiers or \texttt{amod}) and the slot-fillers and their POS tags (e.g., \texttt{new/A}, \texttt{old/A}, \texttt{vertical/A}, \texttt{horizontal/A}). Note that because the CoNLL-U file does not encode the directionality in the \texttt{deprel} column but only in the HEAD indices, we have to use the prefix \texttt{chi\_} (children) if the slot-filler depends on the target and the prefix \texttt{pa\_} (parent) if the target depends on the slot-filler. We preserve distinct dependency labels instead of aggregating semantically similar roles (e.g., combining \texttt{nsubj:pass} and \texttt{obj}). Doing so has two advantages. First, this semantic granularity allows us to detect specific structural shifts, such as an increase in passive voice usage, which would otherwise be obscured by a coarse-grained \texttt{object} category. Second, it is generalisable across languages while merging has to be performed on a case-by-case basis and requires language specific knowledge.

For each target word and for each of its slots, we now have a different distribution of their slot-fillers for each period. To quantify how this distribution changes across periods, we use Jensen-Shannon Divergence (JSD). JSD is a symmetric, smoothed variant of Kullback–Leibler divergence \cite{kullback_information_1951} and when computed with log base 2, it is bounded between 0 and 1, making the scores easy to interpret and comparable across slots. Moreover, JSD can be decomposed into per-filler contributions, enabling fine-grained attribution of which fillers drive a slot’s change. By contrast, prior work on Grammatical Profiles \cite{kutuzov_grammatical_2021, ryzhova_detection_2021} measured change with cosine distance between time-specific frequency vectors, which compares vector geometry (angles) rather than directly comparing probability distributions.

After computing slot-level JSD scores for each target lemma, we aggregate them into a single lemma-level change score by taking the mean JSD over slots whose JSD exceeds 0.5. We use a mean rather than a sum because lemmas differ in how many slots they have and summing would systematically inflate scores for lemmas with more slots. Using the maximum is also undesirable because 1) very rare slots with small sample size can yield extreme JSD values (often 1), which would make many lemmas appear equally ‘maximally changed’ and 2) of our assumption that semantic change is manifested in multiple slots. There is little established guidance on choosing an appropriate JSD threshold, and sparsity becomes more severe when distributions are split by slot, which can push JSD values upward compared to larger distributions. The 0.5 value was empirically selected (based on our previous experience applying JSD to sparse linguistic data) to balance sensitivity to genuine linguistic change against robustness to noise. It serves as a high-pass filter for semantic salience, which isolates substantial structural shifts while suppressing minor fluctuations inherent to sparse distributions. Note that we intentionally avoid any supervised finetuning (e.g., for selecting the optimal JSD cut-off) to preserve the fully unsupervised protocol and comparability with other unsupervised approaches.

To sum up, the inputs to our method are dependency-parsed corpora and a list of target lemmas. SynFlow then
\begin{enumerate}
    \item Collects the slot fillers for each dependency slot;
    \item Computes the JSD scores for every slot;
    \item Aggregates the slot-level JSDs greater than 0.5 into a lemma-level JSD score. 
\end{enumerate} The output is a dictionary of the form \{\texttt{target lemma}: \texttt{aggregate JSD score}\}. In the next sections, we describe 2 improvements to this basic procedure.

\subsection{Frequency filtering}
\label{sec:frequency filtering}
To reduce noise that could be introduced by rare items, we exclude slot-fillers that appear only once in all periods before calculating the JSD.

\subsection{POS removal}
In the basic procedure above, we initially retain POS tags for slot fillers. However, because our analysis is driven primarily by dependency relations, and because POS tags can introduce additional noise (e.g., adjective vs. participial verb ambiguities as discussed above), we remove the POS information from slot-fillers before computing JSD. This reduces the risk that parser-induced POS inconsistencies create spurious differences between periods and ensures that divergence reflects changes in dependency defined co-occurrence rather than tagging artefacts.

\section{Results}

We evaluate our method on both subtasks of the SemEval 2020 Unsupervised Lexical Semantic Change Detection shared task. As described in Section \ref{sec:data and tasks}, Subtask 1 is a binary classification task and is evaluated based on accuracy. Subtask 2 is a ranking task and is evaluated with Spearman’s rank correlation. The main focus of our method would be on \textbf{subtask 2}. We then use our subtask 2 scores for subtask 1 classification, following the strategy of \citet{kutuzov_grammatical_2021}.

\subsection{Subtask 2}

\begin{table*}[t!] 
    \centering
    \small 
    \begin{tabular}{l c c c c c} 
        \toprule
        \textbf{System} & \textbf{Average} & \textbf{English} & \textbf{German} & \textbf{Latin} & \textbf{Swedish} \\ 
        \midrule
        
        \multicolumn{6}{c}{\textbf{Basic Procedure}} \\
        \midrule
        SE & 0.096 & 0.108 & 0.135 & 0.026 & 0.115 \\ 
        SZ & 0.108 & 0.139 & 0.120 & 0.017 & 0.155 \\ 
        \midrule

        \multicolumn{6}{c}{\textbf{Frequency Filtering}} \\
        \midrule
        SE 2 & 0.229 & 0.266 & 0.253 & 0.215 & 0.181 \\ 
        SZ 2 & 0.129 & -0.010 & 0.191 & 0.142 & \textbf{0.192} \\ 
        \midrule

        \multicolumn{6}{c}{\textbf{POS Removal}} \\
        \midrule
        \makecell[l]{SE No POS} & 0.117 & 0.115 & 0.175 & 0.038 & 0.141 \\ 
        \makecell[l]{SZ No POS} & 0.113 & 0.139 & 0.145 & 0.001 & 0.166 \\ 
        \midrule

        \multicolumn{6}{c}{\textbf{Frequency Filtering + POS Removal}} \\
        \midrule
        \makecell[l]{SE 2 No POS} & \textbf{0.239} & \textbf{0.277} & \textbf{0.258} & \textbf{0.258} & 0.162 \\ 
        \makecell[l]{SZ 2 No POS} & 0.139 & -0.008 & 0.193 & 0.197 & 0.173 \\ 
        \midrule

        \multicolumn{6}{c}{\textbf{Prior SemEval results}} \\
        \midrule
        \makecell[l]{Count baseline} & 0.144 & 0.022 & 0.216 & 0.359 & -0.022 \\ 
        \makecell[l]{Best shared task system \\ \cite{cassotti_xl-lexeme_2023}} & 0.583 & 0.757 & 0.877 & -0.056 & 0.754 \\ 
        \bottomrule
    \end{tabular}
    \caption{Performance in graded change detection (SemEval 2020 Subtask 2), Spearman rank correlation coefficients. SE = SemEval, SZ = Stanza.}
    \label{tab:results subtask 2}
    
    \vspace{0.5cm} 
       \begin{tabular}{l c c c c c} 
        \toprule
        \textbf{System} & \textbf{Average} & \textbf{English} & \textbf{German} & \textbf{Latin} & \textbf{Swedish} \\ 
        \midrule
        
        \multicolumn{6}{c}{\textbf{Basic Procedure}} \\
        \midrule
        SE & 0.580 & 0.595 & 0.604 & 0.475 & 0.645 \\ 
        SZ & 0.583 & 0.541 & 0.604 & 0.475 & \textbf{0.71} \\ 
        \midrule

        \multicolumn{6}{c}{\textbf{Frequency Filtering}} \\
        \midrule
        SE 2 & 0.593 & \textbf{0.649} & 0.604 & 0.475 & 0.645 \\ 
        SZ 2 & 0.569 & 0.486 & 0.604 & 0.475 & \textbf{0.71} \\ 
        \midrule

        \multicolumn{6}{c}{\textbf{POS Removal}} \\
        \midrule
        \makecell[l]{SE No POS} & 0.59 & 0.595 & \textbf{0.646} & 0.475 & 0.645 \\ 
        \makecell[l]{SZ No POS} & 0.583 & 0.541 & 0.604 & 0.475 & \textbf{0.71} \\ 
        \midrule

        \multicolumn{6}{c}{\textbf{Frequency Filtering + POS Removal}} \\
        \midrule
        \makecell[l]{SE 2 No POS} & \textbf{0.606} & \textbf{0.649} & 0.604 & \textbf{0.525} & 0.645 \\ 
        \makecell[l]{SE 2 No POS DP} & 0.585 & 0.621 & 0.604 & 0.5 & 0.613 \\ 
        \makecell[l]{SZ 2 No POS} & 0.565 & 0.486 & 0.604 & \textbf{0.525} & 0.645 \\ 
        \midrule

        \multicolumn{6}{c}{\textbf{Prior SemEval results}} \\
        \midrule
        \makecell[l]{Count baseline} & 0.613 & 0.595 & 0.688 & 0.525 & 0.645 \\ 
        \makecell[l]{Best shared task system} & 0.687 & 0.622 & 0.75 & 0.7 & 0.677 \\ 
        \bottomrule
    \end{tabular}
    \caption{Performance in binary change detection (SemEval 2020 Subtask 1), accuracy. Note that in this paper we mostly focus on ranking (Subtask 2). All the binary change detection methods here are entirely based on the scores produced by the ranking methods. SE = SemEval, SZ = Stanza.}
    \label{tab:results subtask 1}
\end{table*}

The results for subtask 2 are reported in table \ref{tab:results subtask 2}. For a full comparison with SemEval-2020 Task 1 participating systems, see Table~5 in \citet{schlechtweg_semeval-2020_2020}.

\paragraph{Basic procedure} The slot-level JSD aggregation described above yields an average Spearman correlation of 0.096 on the SE dataset and 0.108 on the SZ dataset. This is below the shared-task baseline, but still higher than some systems based on type and token embeddings (e.g., UoB, TUE).

\paragraph{Frequency filtering} Removing rare slot-fillers (Section \ref{sec:frequency filtering}) improves the performance substantially, especially on the SE dataset: the average correlation across languages nearly doubles, and Latin improves by an order of magnitude ($0.026 \to 0.215$). On the SZ dataset, the effect is mixed: English drops to $-0.001$, while the other three languages improve.

\paragraph{POS Removal} Removing POS tags from slot-fillers yields only a small additional gain: $+0.021$ on SE and $+0.005$ on SZ on average.

\paragraph{Frequency Filtering + POS Removal} Combining both strategies produces our strongest overall results. Performance improves for all languages except Swedish (where the combined setting reaches $0.162$, slightly below our Swedish best of $0.192$). The combined setting on the SE dataset achieves an average Spearman correlation of $0.239$, which is nearly double the Count baseline and higher than a number of embedding-based systems. This is impressive given the simplicity of the approach.

\subsection{Subtask 1}
The results for subtask 1 are presented in table \ref{tab:results subtask 1}. We follow the thresholding strategy used by \citet{kutuzov_grammatical_2021} by assigning a classification score of 1 to the top $43\%$ of the target words for each language. The ranking is obtained from their JSD scores as discussed above. The different average accuracy scores from the different systems are tightly clustered (between 0.58 and 0.606) and are very close to the accuracy of the baseline, which is already surprisingly hard to beat.

Our best system is still \texttt{SE 2 No POS} (i.e., SemEval + frequency filtering + POS removal), which achieves an average accuracy of $0.606$, showing that simple denoising techniques (i.e., frequency filtering and POS removal) work. In the case of Swedish, 3 of our system variants even outperform the state-of-the-art, achieving an accuracy of $0.71$. We also tested our best system with dynamic programming\footnote{We applied the offline change point detection algorithm on the sorted JSD scores. The method identifies a split point $k$ that minimises the total within-segment squared error: $\text{SSE}(y_{1:k}) + \text{SSE}(y_{k+1:n})$, where $\text{SSE} = \sum y^2 - \frac{(\sum y)^2}{n}$.} \cite{truong_selective_2020} and observed that it does not cause any significant accuracy decrease when compared to the hard-coded $43\%$ cut-off point. This indicates that our method does not require knowledge of the test data distribution

\section{Qualitative analysis}
\label{sec:qualitative analysis}

The strong point of our system is its interpretability, in the sense that it allows the users to see which kinds of shift drive the change through JSD decomposition, and check the concordances for the locus of semantic change. In this section, we analyse the prediction of the system qualitatively to understand what is being captured.

Based on the rankings of the Subtask 2 Gold Scores and our system predictions, we divided the target lemmas into five groups: True Positive (TP), True Negative (TN), False Positive (FP), False Negative (FN) and MID. A lemma is TP if it falls in the top $33\%$ of both the gold and predicted rankings, and TN if it falls in the bottom $33\%$ of both. A lemma is FP if it is in the bottom $33\%$ of the gold ranking but the top $33\%$ of our method's ranking, and FN if it is in the top $33\%$ of the gold ranking but the bottom $33\%$ of our method's ranking. All remaining lemmas are assigned to the MID group. We focus our qualitative analysis on TP, TN, FP, and FN, as these categories most clearly reveal our method's strengths and limitations. Due to space constraints, in the main text we present one illustrative example per group and language (the most extreme cases). The lemmas of each group are listed in Table~\ref{tab:appendix_lemmas} in the Appendix.

Also, in the analysis, we will only look at slots with JSD of at least $0.5$ as that is the cut-off point used in the previous procedure. A detailed description of the dependency slots presented in this section is provided in Table~\ref{tab:appendix_slots} in the Appendix. 

\begin{figure}[b!]
    \centering
    \includegraphics[width=\linewidth]{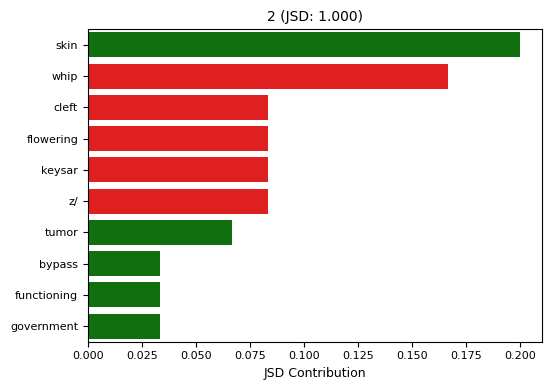}
    
    \caption{The JSD contributions of different slot-fillers in the slot \texttt{chi\_compound} of the English noun \textit{graft}. Green bars indicate an increase in relative frequency; red bars indicate a decrease.}
    \label{fig:graft}
\end{figure}

\paragraph{True positives} Upon qualitative analysis, our method does indeed capture relevant co-occurrence information related to semantic change. The English noun \textit{graft} develops a new \textsc{medical} use (transplanting tissue) and a new \textsc{political} use (political corruption) in the later period, whereas in the earlier period it is used primarily in agricultural contexts. This is reflected in 3 out of 5 dependency slots, such as governing verbs (\texttt{pa\_obj}, e.g., \textit{cut the graft}) and compound modifiers (\texttt{chi\_compound}, e.g., \textit{skin graft}) as shown in Figure~\ref{fig:graft}. In the first period, fillers and contexts predominantly relate to cultivation and horticulture (e.g., \textit{adapting the graft to the soil}, \textit{flowering graft}), while the second period is characterised by an increase of medical and political contexts (e.g., \textit{help the skin graft}, \textit{government graft}). The remaining two slots show weaker or less interpretable shifts. The German adjective \textit{abgebrüht} has developed a new \textsc{figurative} meaning (e.g., \textit{abgebrühter Manager} `a hardened manager') from the \textsc{literal} one (e.g., \textit{abgebrühter Reis} `blanched rice'), as evidenced in the modified nouns (\texttt{pa\_amod}) and modifying adverbs (\texttt{chi\_advmod}). The Latin noun \textit{pontifex} shifts from `state-religious officer' toward `the Pope'. This is reflected in 13/15 slots. In the adjectival modifier (\texttt{chi\_amod}) and oblique nominals (\texttt{pa\_obl}) slots, for example, \textit{pontifex} co-occurs less with administrative vocabulary (e.g., \textit{gero} `manage') and more with Christian contexts (e.g., \textit{summus pontifex} `Supreme Pontiff', \textit{oro} `pray'). Finally, the Swedish noun \textit{uppläggning} has developed a new abstract \textsc{structuring} meaning from the physical \textsc{stockpiling} meaning, with consistent evidence in all of its slots. For example, in the nominal modifier slot (\texttt{pa\_obl}), there is an increase in \textit{räkenskap} `accounting' while in the earlier period it is dominated by types of goods such as \textit{jernplätars} `of iron plates', \textit{spannmåls} `of cereals'.

\paragraph{True negatives} Regarding TNs, we observe that few dependency slots exceed the JSD threshold and even when they do, the margin is often negligible. This is the case for the English noun \textit{tree}. Our method detects a moderate shift in nominal modifiers (\texttt{chi\_nmod}, JSD $= 0.517$) and compounds (JSD $= 0.533$), but these shifts are driven by a transition from biological attributes (e.g., \textit{trees of this size}) to locative contexts (e.g., \textit{trees of this yard}). Crucially, however, both filler sets remain semantically consistent with the core \textsc{plant} sense, indicating mere topical variation rather than genuine semantic evolution. Even if there are slots with high divergence (e.g., $\approx 0.8$), the aggregate lemma-level score remains low because other stable slots dilute the overall mean. This was the case for the German verb \textit{vergönnen}, the Latin adjective \textit{necessarius} and the Swedish verb \textit{uträtta}.

\paragraph{False positives} FPs are target words for which our method detects large shifts in one or more slots, yet the lemma is ranked low in the gold Subtask 2 ranking. A common pattern is slot-level turnover driven by a small number of highly characteristic contexts, which inflates divergence without corresponding to a clear sense innovation under the annotation scheme. For the English noun \textit{face}, the high change score is driven primarily by the two compound slots (\texttt{pa\_compound}, \texttt{chi\_compound}). The divergence is dominated by a few fixed expressions, most notably \textit{face value} and product-related compounds such as \textit{face cream} and \textit{face lift}. This illustrates the limitation where the substantial frequency change of a small set of expressions can cause the method to overestimate the semantic change. The Swedish noun \textit{kokärt} is also stable in meaning, yet all of its slots exhibit high JSD due to contextual turnover. The German noun \textit{Seminar} presents an interesting case. We observe a metonymic shift: \textsc{institutional department} $\to$ \textsc{courses}. This is strongly evidenced by a high divergence ($0.926$) in the nominal heads (\texttt{pa\_nmod}), transitioning from institution-related fillers (e.g., \textit{Lehrer}, \textit{Director}, \textit{Übungsschule}) to participant-oriented ones (e.g., \textit{Teilnehmer}). However, its official gold label for Subtask 1 is 0 (i.e., no sense gain or loss) and its official Subtask 2 rank is only 39/48 (i.e., low degree of semantic change). This suggests our system successfully detected an overlooked semantic shift in the official annotations. The Latin noun \textit{dolus} is another interesting case. It is officially classified as $1$ in Subtask 1 and our method successfully identified the word's narrowing and pejoration from \textsc{trick} (neutral) to \textsc{fraud} (negative) in the governing verb (\texttt{pa\_obj}) slot (JSD $= 0.933$). However, its official ranking in Subtask 2 is low (29/40). This strengthens the organisers' finding: models that excel at one subtask are not necessarily good at the other \cite{schlechtweg_semeval-2020_2020}.

\paragraph{False negatives} FNs are target words whose aggregate JSD scores underestimate the degree of change with respect to the gold ranking. We attribute a substantial portion of these errors to our slot-selection threshold. Although the English noun \textit{record} develops a salient \textsc{best performance} sense (e.g., \textit{world record}), reflected by a sharp rise of \textit{set} and \textit{break} in governing verbs (\texttt{pa\_obj}), the divergence for this slot is $0.48$, just below our $0.5$ cutoff. This signal is thus excluded from aggregation, causing the lemma-level score to miss the underlying shift. A similar threshold effect appears for the Swedish noun \textit{motiv}, which shifts from a predominantly \textsc{reason} meaning (typically framed in moral evaluation) to a predominantly \textsc{artistic subject} meaning (often framed in aesthetic evaluation) between the two periods. While detectable in adjectival modifiers (\texttt{chi\_amod}), the slot's low JSD ($0.436$) leads to its exclusion, again resulting in an underestimated aggregate score. In other cases, the change signal is present but remains only moderately strong at the slot level, so it does not lift the aggregate score enough to match the gold ranking. The German verb \textit{ausspannen} shows a clear metaphorisation: \textsc{physical} (unharnessing/stretching out) $\to$ \textsc{relax} (taking time off). This is visible in its direct objects (\texttt{chi\_obj}) with an increase in temporal fillers (\textit{Tag}, \textit{Stunde}) and a decrease in physical objects (\textit{Draht}, \textit{Faden}), or in the passive subject (\texttt{chi\_nsubj:pass}). However, their JSDs are only $0.543$ and $0.584$, respectively. This illustrates that genuine semantic innovations can yield relatively modest divergences in individual slots, resulting in underestimated lemma-level change. A related pattern occurs for the Latin noun \textit{potestas}, where broadening (\textsc{legal authority} $\to$ \textsc{general capability}) is reflected in two low JSD slots, adjectival modifiers (\texttt{chi\_amod}, JSD $= 0.587$) and nominal modifiers (\texttt{chi\_nmod}, JSD $= 0.643$).

\section{Conclusion}

This paper demonstrates that explicit linguistic co-occurrence evidence, when structured through syntax, is sufficient to build an effective lexical semantic change detection system. Our method achieves competitive performance across languages and can even outperform several distributional models in some tasks and languages. Although our method does not match state-of-the-art systems based on large pretrained language models, which can encode a wider range of linguistic and contextual information, maximising performance metrics was not our primary objective. Instead, we want to bring more attention to theory-driven and interpretable approaches.

To further refine this approach, future work should:

\begin{enumerate}
    \item Replace hard slot thresholds with more adaptive, evidence-weighted aggregation to increase reliability; 
    \item Further develop theory-driven ways of selecting and merging slots to reduce non-essential variations (recall the \texttt{nsubj:pass} and \texttt{obj} example in \ref{sec:basic procedure}), but note the cost of generalisability;
    \item Explore hybrid methods that preserve slot-level attribution while using embeddings to stabilise sparse evidence (e.g., clustering slot-fillers using general type embeddings before computing JSD);
    \item Take into account covariations between slots and looking at multiple slots simultaneously (e.g., to detect changes in valency/subcategorisation patterns)
\end{enumerate}

More broadly, we view our method as an instance of a general framework for semantic change analysis: identify interpretable dimensions of a target's distribution (here, dependency relations) and quantify change within each dimension before aggregating into a global measure. This framework could (and should) be used as a complementary tool to interpret and validate the changes detected by other approaches.

\section{Limitations}

We highlight three primary limitations and practical considerations of our proposed method. 

First, dependency availability is a hard constraint, making this method inapplicable to low-resource languages lacking robust dependency parsers. Even for high-resource languages, pre-processing choices matter: POS and lemma mismatches between gold targets and parsers (e.g., Stanza) output requires alignments. 

Second, sensitivity to noise (e.g., rare slots, rare slot-fillers) remains a challenge. Denoising (e.g., filtering singletons and removing POS tags from fillers) proved central to our performance. 

Third, our qualitative analysis, while supporting the interpretability claim, simultaneously reveals inherent constraints in our scoring mechanism. The method can \textit{overestimate} change when a small number of fixed expressions or highly productive co-occurrences dominate slot turnover (as seen with idioms). Conversely, it can \textit{underestimate} change when genuine semantic innovations fall just below the heuristic slot-selection cutoff.

\section*{Acknowledgments}
This project has received funding from the European Union’s Horizon Europe programme for research and innovation under MSCA Doctoral Networks 2022, Grant Agreement No. 101120349 and Grant Agreement No. 101119511.

We would like to thank the anonymous reviewers for their detailed and informative comments and suggestions.
\bibliography{custom}
\bibliographystyle{acl_natbib}

\appendix

\onecolumn 
\appendix
\section*{Appendix}
\section{Predictions of our method}
\label{sec:appendix_predictions}

\begin{table}[H] 
    \centering
    \small
    \begin{tabular}{l l l l l} 
        \toprule
        & \textbf{English} & \textbf{German} & \textbf{Latin} & \textbf{Swedish} \\ 
        \midrule
        \textbf{True Positive} & 
        \makecell[tl]{graft (noun) \\ plane (noun) \\ head (noun) \\ bit (noun) \\ player (noun)} & 
        \makecell[tl]{abgebrüht \\ Engpaß \\ Ohrwurm \\ Abgesang \\ Eintagsfliege \\ Dynamik \\ Armenhaus} & 
        \makecell[tl]{pontifex \\ sacramentum \\ scriptura \\ titulus \\ imperator \\ sanctus} & 
        \makecell[tl]{uppläggning \\ krita \\ medium \\ konduktör \\ uppfattning \\ beredning \\ notis} \\ 
        \midrule
        \textbf{True Negative} & 
        \makecell[tl]{tree (noun) \\ risk (noun) \\ multitude (noun) \\ fiction (noun) \\ contemplation (noun) \\ bag (noun)} & 
        \makecell[tl]{vergönnen \\ Tier \\ Ackergerät \\ Frechheit \\ aufrechterhalten} & 
        \makecell[tl]{necessarius \\ consilium \\ hostis \\ simplex \\ voluntas \\ poena} & 
        \makecell[tl]{uträtta \\ annandag \\ studie} \\ 
        \midrule
        \textbf{False Positive} & 
        \makecell[tl]{face (noun) \\ quilt (noun) \\ lane (noun) \\ savage (noun)} & 
        \makecell[tl]{Seminar \\ Truppenteil \\ Mulatte \\ Pachtzins \\ Naturschönheit} & 
        \makecell[tl]{dolus \\ fidelis \\ consul \\ ancilla \\ acerbus} & 
        \makecell[tl]{kokärt \\ vegetation} \\ 
        \midrule
        \textbf{False Negative} & 
        \makecell[tl]{record (noun) \\ ounce (noun) \\ rag (noun)} & 
        \makecell[tl]{ausspannen \\ artikulieren \\ verbauen} & 
        \makecell[tl]{potestas \\ licet \\ salus \\ virtus} & 
        \makecell[tl]{motiv \\ ledning \\ bearbeta} \\ 
        \bottomrule
    \end{tabular}
    \caption{True Positive, True Negative, False Positive, False Negative results of \textsc{SE 2 NoPOS}.}
    \label{tab:appendix_lemmas}
\end{table}

\section{Descriptions of the slots}
\label{sec:appendix_slots}

\begin{table}[H] 
    \centering
    \small
    \begin{tabular}{l l} 
        \toprule
        \textbf{Slot Name} & \textbf{Explanation} \\ 
        \midrule
        \texttt{chi\_advmod} & The target is modified by the adverbial slot filler. \\ 
        \texttt{chi\_amod} & The target is modified by the adjectival slot-filler. \\ 
        \texttt{chi\_case} & The target is governing the prepositional slot-filler. \\ 
        \texttt{chi\_compound} & The target is being modified by the slot-filler in the compound. \\ 
        \texttt{chi\_nmod} & The target is modified by the nominal slot-filler. \\ 
        \texttt{chi\_nsubj:pass} & The target is the verb governing the passive subject. \\ 
        \texttt{chi\_obj} & The target is the verb governing the object slot-filler. \\ 
        \texttt{pa\_amod} & The target is the adjective modifying the slot-filler. \\ 
        \texttt{pa\_compound} & The target modifies the slot-filler in the compound. \\ 
        \texttt{pa\_nmod} & The target is the nominal modifier of the slot-filler. \\ 
        \texttt{pa\_obj} & The target is the object of the verbal slot-filler. \\ 
        \texttt{pa\_obl} & The target is the oblique of the slot-filler. \\ 
        \bottomrule
    \end{tabular}
    \caption{A detailed description of the dependency slots presented in section \ref{sec:qualitative analysis}.}
    \label{tab:appendix_slots}
\end{table}
\twocolumn

\end{document}